\documentclass{research4cacm}
\usepackage{amsmath}
\usepackage{mathtools}
\usepackage{multirow}
\usepackage{xcolor}
\usepackage{makecell}
\usepackage{graphics}
\usepackage{adjustbox}
\usepackage{enumitem,booktabs,cfr-lm}
\usepackage{amssymb}
\usepackage[ruled,linesnumbered]{algorithm2e}
\usepackage{url}
\usepackage{diagbox}
\usepackage{balance}

\begin{document}

\title{Evaluating Explanation Without Ground Truth in Interpretable Machine Learning}

\numberofauthors{3} 

\author{
Fan Yang, Mengnan Du, Xia Hu \\
Department of Computer Science and Engineering \\
Texas A\&M University \\
\{nacoyang, dumengnan, xiahu\}@tamu.edu
}

\maketitle
\begin{abstract}
Interpretable Machine Learning (IML) has become increasingly important in many real-world applications, such as autonomous cars and medical diagnosis, where explanations are significantly preferred to help people better understand how machine learning systems work and further enhance their trust towards systems. However, due to the diversified scenarios and subjective nature of explanations, we rarely have the ground truth for benchmark evaluation in IML on the quality of generated explanations. Having a sense of explanation quality not only matters for assessing system boundaries, but also helps to realize the true benefits to human users in practical settings. To benchmark the evaluation in IML, in this article, we rigorously define the problem of evaluating explanations, and systematically review the existing efforts from state-of-the-arts. Specifically, we summarize three general aspects of explanation (i.e., \textit{generalizability}, \textit{fidelity} and \textit{persuasibility}) with formal definitions, and respectively review the representative methodologies for each of them under different tasks. Further, a unified evaluation framework is designed according to the hierarchical needs from developers and end-users, which could be easily adopted for different scenarios in practice. In the end, open problems are discussed, and several limitations of current evaluation techniques are raised for future explorations. 
\end{abstract}

\section{Introduction}

\begin{figure}[t]
  \centering
  \includegraphics[width=\columnwidth]{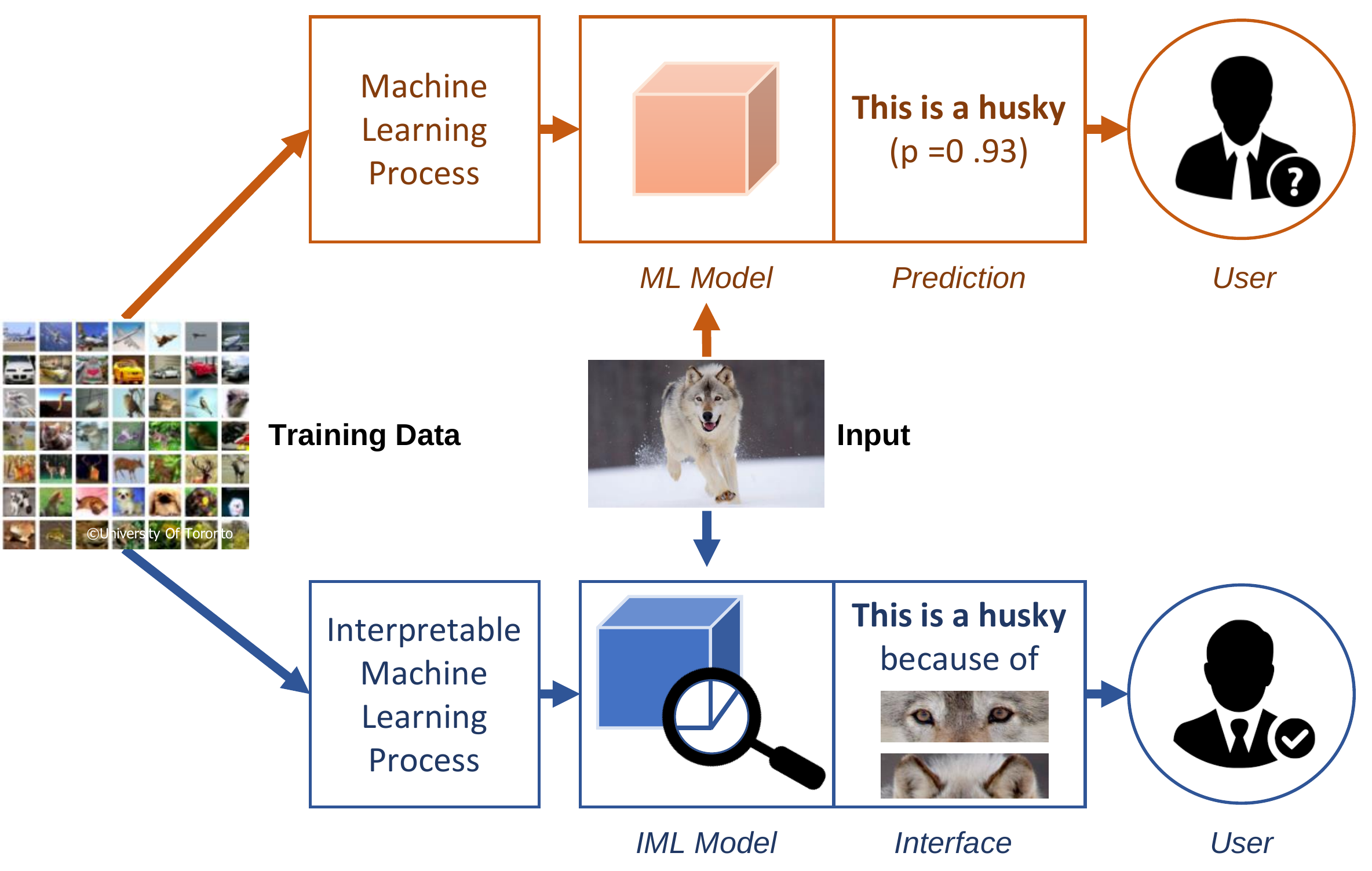}
  \caption{Illustration of the IML techniques. We compare the two different pipelines between machine learning (ML) and IML. It is worth noting that IML model is capable of providing specific reasons for particular machine decisions, while ML model may simply provide the prediction results with probability scores. Here, we employ the image classification task as an example, where IML model could tell which part of the image contributes the animal to a husky while ML model may only tell the overall confidence towards a husky classification result.}
  \label{fig:imlintro}
\end{figure}

Serving as one of the most significant momentums for the booming of artificial intelligence, machine learning is playing a vital role in many real-world systems, widely ranging from spam filters to humanoid robot. To handle the tasks that are increasingly complicated in practice nowadays, more and more sophisticated machine learning systems are designed, such as deep learning models~\cite{lecun2015deep}, for accurate decision making. Despite the superior performance, those complex systems are typically hard to be interpreted by human users, which largely limits their applications in some high-stake scenarios like self-driving vehicles and medical treatment, where explanations are important and necessary for scrutable decisions~\cite{wachter2017transparent}. To this end, the concept of interpretable machine learning (IML) has been formally raised~\cite{doshi2017towards}, aiming to help humans better understand the machine learning decisions. We illustrate the core idea of IML techniques in Figure~\ref{fig:imlintro}.

\begin{figure}[t]
  \centering
  \includegraphics[width=\columnwidth]{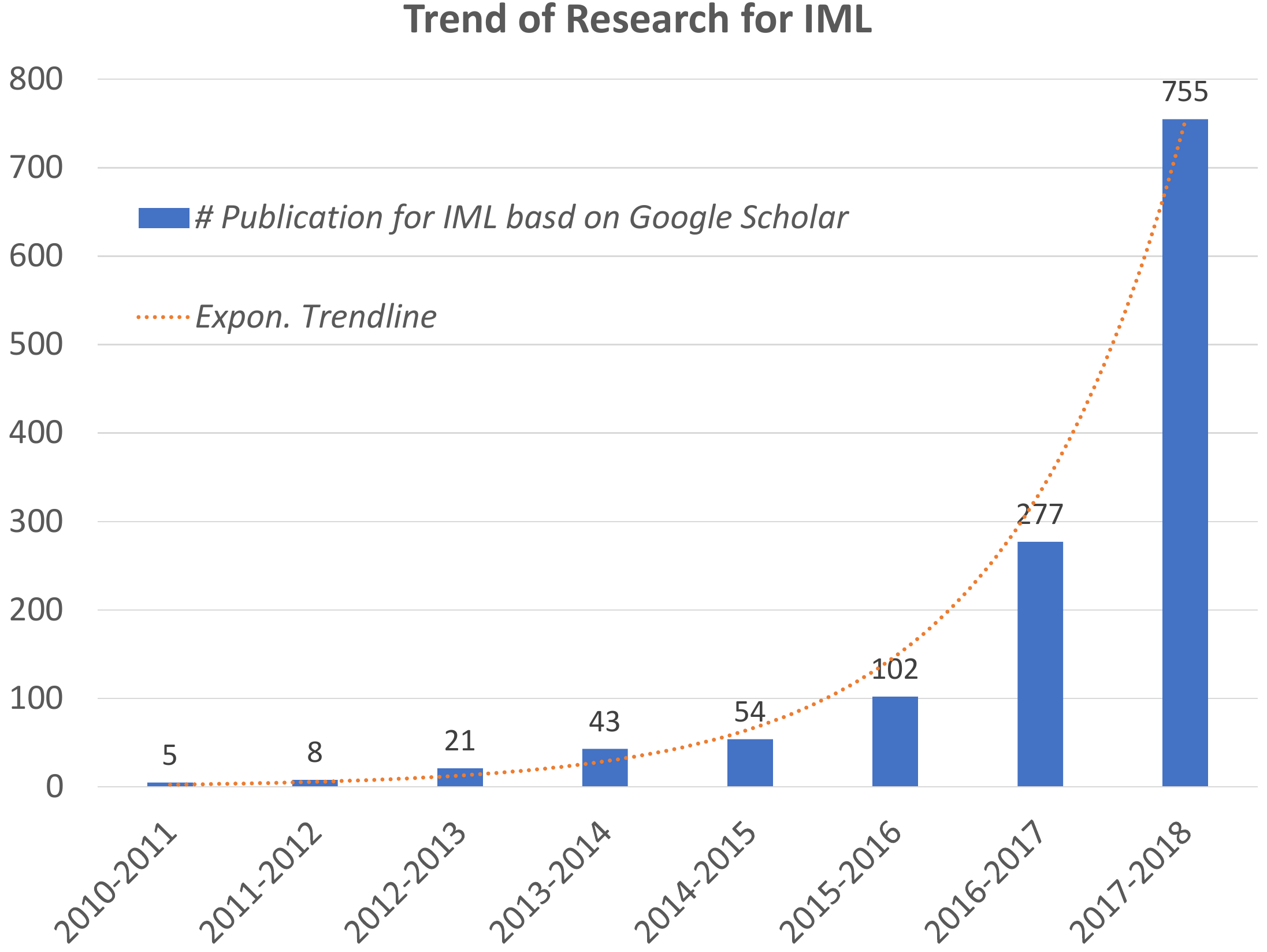}
  \caption{Tendency of the IML research in recent years. In particular, we present the number of research publications related to IML from 2010 to 2018, and plot the trendline according to the statistics. The relevant numerics are collected from Google Scholar, with the key words ``interpretable machine learning". We believe the actual numbers are even larger than the provided, since some other terms, such as ``explainable", which are closely related to IML, are ignored during collection. From the results, we can see that IML related publication has been increasing exponentially, and much more attention has been paid for this field.}
  \label{fig:imlnum}
\end{figure}

IML is a new branch of machine learning techniques with mounting attentions in recent years (shown by Figure~\ref{fig:imlnum}), focusing on the decision explanation beyond the instance prediction. IML is typically employed to extract useful information, from either system structure or prediction result, as explanations to interpret relevant decisions. Although IML techniques have been comprehensively discussed covering methodology and application~\cite{du2018techniques}, the insights on IML evaluation perspective are still rather limited, which significantly impedes the way of IML to a rigorous science field. To precisely reflect the boundaries of IML systems and measure the benefits of explanations brought to human users, effective evaluations are pivotal and indispensable. Different from the conventional evaluation purely relied on model performance, IML evaluation also needs to pay attention to the quality of the generated explanations, which makes it hard to be handled and benchmarked. 

Evaluating explanation in IML is a challenging problem, since we need to balance well between the \emph{objective} and \emph{subjective} perspectives when designing experiments. On one hand, different users could have different preferences towards what a good explanation should be under different scenarios~\cite{doshi2017towards}, thus it is not practical to benchmark the IML evaluation with a common set of ground truth for objective evaluations. For example, when deploying self-driving cars with IML, system engineers may consider sophisticated explanations as good ones for safety concerns, while car drivers may prefer those concise explanations because complex ones could be too time-consuming for decision making during driving. On the other hand, there might be more criterion beyond human subjective satisfaction. Human preferred explanations may not always represent the full working mechanism of systems, which could lead to a poor performance on generalization. It has shown that subjective satisfaction of explanations largely depends on the response time of human users, and has no clear relation with the accuracy performance~\cite{lage2019evaluation}. This finding directly supports the fact that human satisfaction cannot be regarded as the sole standard when evaluating explanation. Besides, fully subjective evaluations would also result in ethics issues, because it is unmoral to manipulate an explanation to better cater human users~\cite{herman2017promise}. Seeking human satisfaction excessively could cause explanations to persuade users, instead of actually interpreting systems. 

Considering the aforementioned challenges, we aim to pave the way for benchmark evaluation in this article, regarding to the explanation generated from IML techniques. First, we give an overview about the explanations in IML, and categorize them by a two-dimensional standard (i.e., interpretation \emph{scope} and interpretation \emph{manner}) with representative examples. Then, we summarize three general properties (i.e., \emph{generalizability}, \emph{fidelity} and \emph{persuasibility}) for explanation with formal definitions, and rigorously define the problem of evaluating explanation in the IML context. Next, following those properties, we conduct a systematic review about existing work on explanation evaluation, with the focus on different techniques in various applications. Moreover, we also review some other special properties for explanation evaluation, which are typically considered under particular scenarios. Further, with the aid of those general properties, we design a unified evaluation framework aligned with the hierarchical needs from both model developers and end-users. At last, we raise several open problems for current evaluation techniques, and discuss some potential limitations for future exploration.

\section{Explanation and Evaluation}

In this section, we first introduce the explanations we particularly focus on, and give an overview about the categories of explanations in IML. Then, three general properties of explanation are summarized for evaluation tasks according to different perspectives in nature. Finally, we formally define the problem of evaluating explanations in IML with the aid of those general properties. 

\subsection{Explanation Overview}

\begin{figure}[t]
  \centering
  \includegraphics[width=\columnwidth]{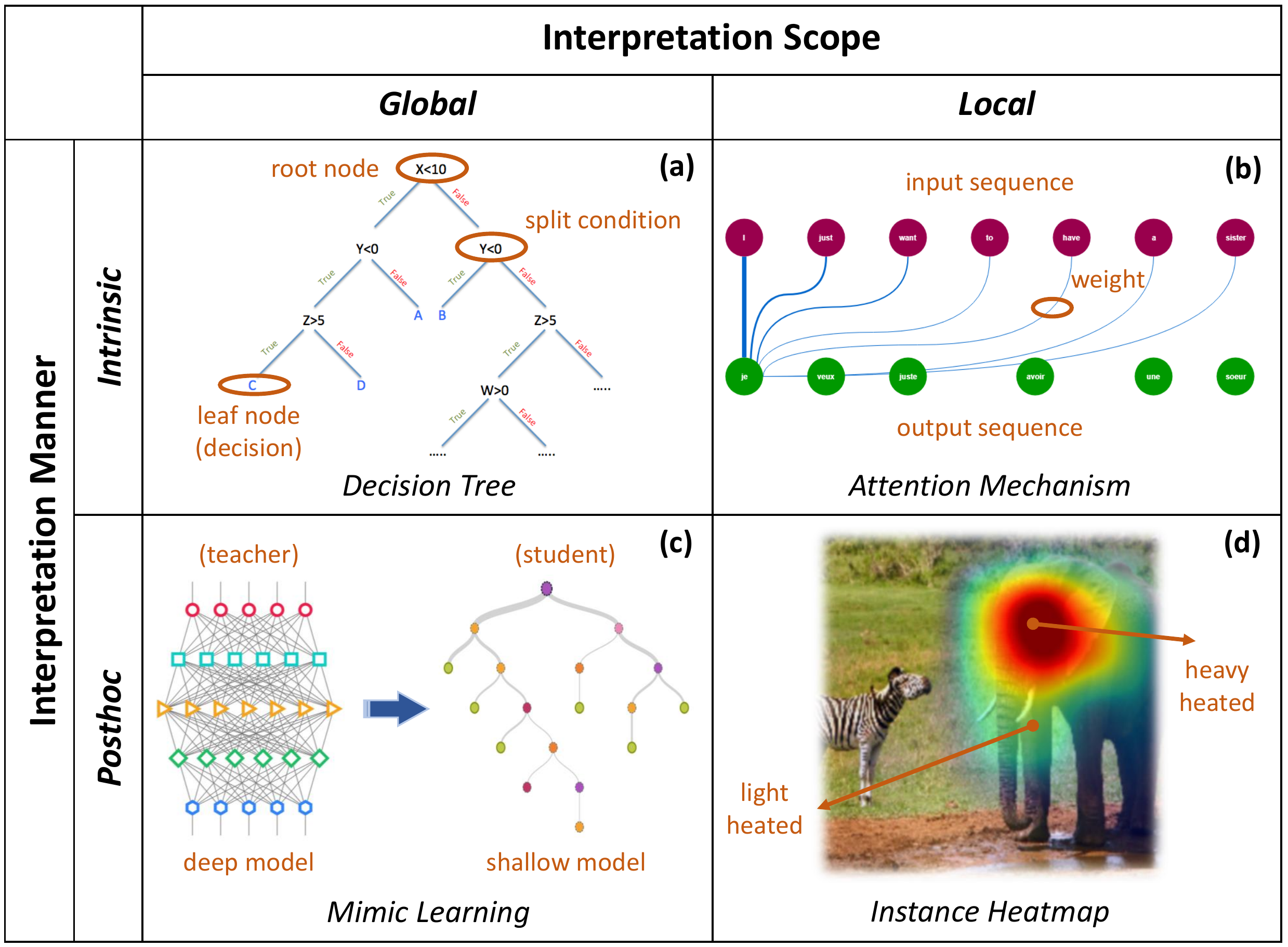}
  \caption{A two-dimensional categorization for explanations in IML, covering interpretation scope and interpretation manner. According to the two-dimensional standard, we can divide explanations into four different groups: (a) \emph{intrinsic-global}; (b) \emph{intrinsic-local}; (c) \emph{posthoc-global}; (d) \emph{posthoc-local}. For each category, we attach a representative example for illustration. In particular, we employ decision tree as the example for intrinsic-global explanations, attention mechanism for intrinsic-local ones, mimic learning for posthoc-global ones, and instance heatmap for posthoc-local ones.} 
  \label{fig:exptype}
\end{figure}

In the context of IML, explanations are particularly referred to those information that can help human users interpret either learning process or prediction result for machine learning models. With different focuses, explanations in IML could have diversified forms with various characteristics, such as the local heatmaps for instances and the decision rules for models. In this article, we specifically categorize the explanations in IML with a two-dimensional standard, covering both interpretation scope and interpretation manner. As for the scope dimension, explanations can be classified into the \emph{global} and \emph{local} ones, where global explanation indicates the overall working mechanism of models by interpreting structures or parameters, and local explanation reflects the particular model behaviour for individual instance by analyzing specific decisions. Regarding to the manner dimension, we can divide explanations into the \emph{intrinsic} and \emph{posthoc} (also written as post-hoc or post hoc) ones. Intrinsic explanation is typically achieved by those self-interpretable models that are transparent with particular designs, while posthoc one requires another independent interpretation model or technique to provide explanations for the target model. The two-dimensional taxonomy of explanations in IML is illustrated by Figure~\ref{fig:exptype}. 

The first category is \emph{intrinsic-global} explanation. This type of explanation can be well represented by some conventional machine learning techniques, such as rule-based systems and decision trees, which are self-interpretable and capable of showing the overall working patterns for prediction. Take the decision tree for example, the intuitional structure, as well as the set of all decision branches, constitutes the corresponding intrinsic-global explanation. The second category is \emph{intrinsic-local} explanation, which is associated with specific input instances. A typical example is the attention mechanism applied on sequential models, where generated attention weights can help interpret particular predictions by indicating the important components. Attention model is widely used in both image captioning and machine translation tasks. \emph{Posthoc-global} explanation serves as the third category, and the representative example can be shown with mimic learning techniques for deep models. As for mimic learning, the teacher usually is a deep model, while the student is typically deployed as a shallow model that is easier to be interpreted. The overall process of mimic learning can be regarded as a distillation process from the teacher to the student, where the interpretable student model provides a global view in a posthoc manner for the deep teacher model. The \emph{posthoc-local} explanation fills up the last part of the taxonomy. We introduce this category with an example of instance heatmap, which is used to visualize the input regions with attribution score (i.e., a quantified importance indicator). Instance heatmap works well for both image and text, and is capable of showing the local behaviour of the target model. Since heatmap depends on the particular input and does not involve the specific model design, it is a typical local explanation within a posthoc way.

\subsection{General Properties of Explanation} 

\begin{figure}[htbp]
  \centering
  \includegraphics[width=\columnwidth]{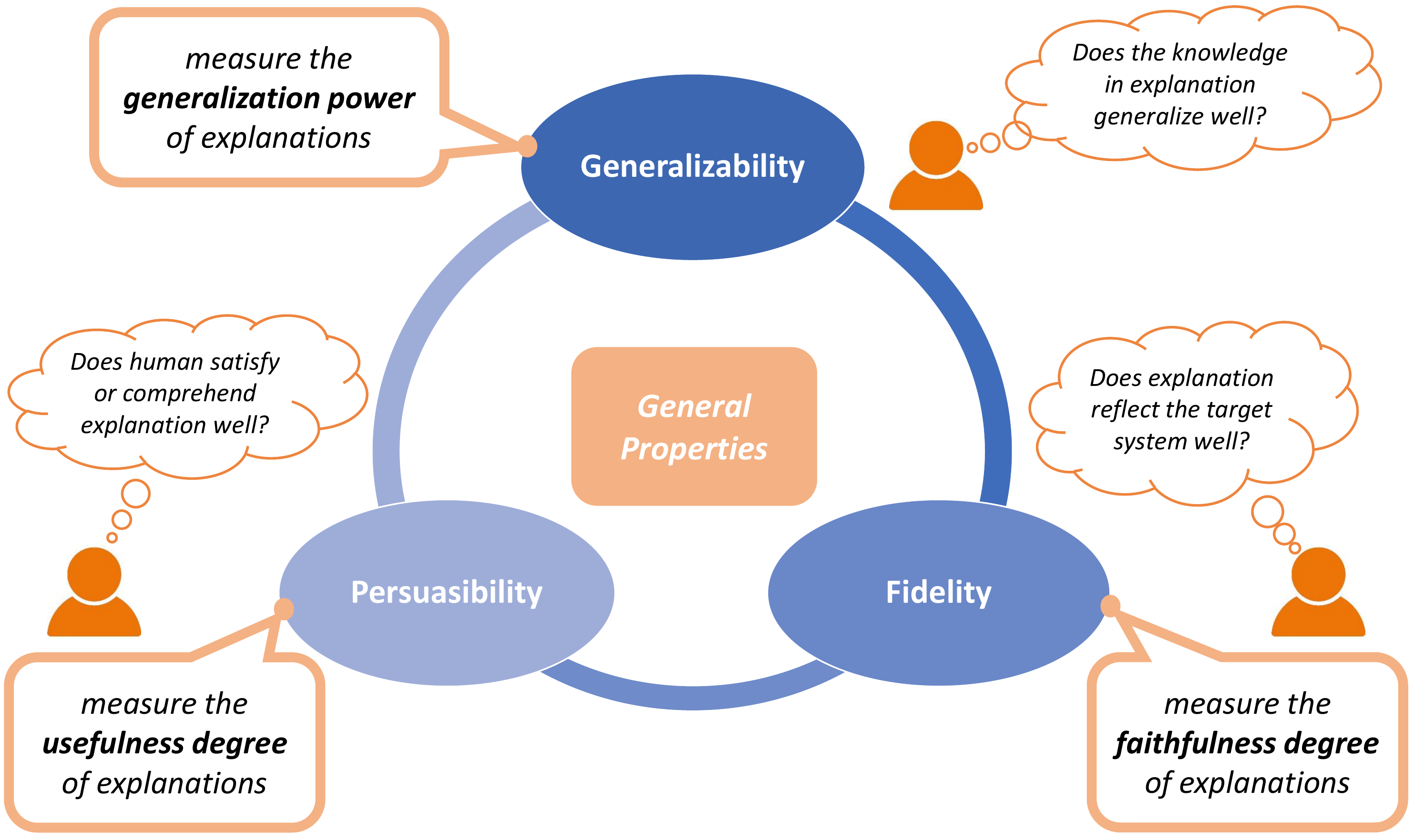}
  \caption{Three general properties for explanations in IML, including generalizability, fidelity and persuasibility. Each property essentially corresponds to one specific aspect in evaluation. Generalizability focuses on the generalization power of explanation. Fidelity focuses on the faithfulness degree of explanation. Persuasibility focuses on the usefulness degree of explanation. }
  \label{fig:evalprop}
\end{figure}

To formally define the problem of evaluating explanations in IML, it is important to make clear the general properties of explanation for evaluation. In this article, we summarize three significant properties from different perspectives, i.e., \textit{generalizability}, \emph{fidelity} and \emph{persuasibility}, where each of the property corresponds to one specific aspect in evaluation. The intuitions of the properties are illustrated in Figure~\ref{fig:evalprop}.

The first general property is \emph{generalizability}, which is used to reflect the generalization power of explanation. In real-world applications, human users employ explanation from IML techniques mainly to obtain insights from the target system, which naturally brings forward the demand on explanation generalization performance. If a set of explanations is poorly generalized, it can hardly be regarded with good quality, since the knowledge and guidance it provides would be rather limited in practice. One thing to clarify is that the explanation generalization mentioned here is not necessarily equal to the model predictive power, unless the model itself is interpretable with self-explanations (e.g., decision tree). By measuring the generalizability of explanation, users can have a sense of how accurate the generated explanations are for specific tasks. \\

\noindent
\fbox{
  \parbox{0.452\textwidth}{
  \textbf{Definition 1:} We define the \emph{generalizability} of explanation in IML as an indicator for generalization performance, regarding to the knowledge and guidance delivered by the corresponding explanation.}
}  \\

The second general property is \emph{fidelity}, which is used to indicate how faithful explanations are to the target system. Faithful explanation is always preferred by human, because it can precisely capture the decision making process of the target system and show the correct evidences for particular predictions. Explanations with high quality need to be faithful, since they are essentially served as important tools for users to understand the target system. Without sufficient fidelity, explanations can only provide limited insights to the system, which degrades the functionalities of IML to human users. To guarantee the relevance of explanations, we need fidelity to conduct explanation evaluation in IML. \\

\noindent
\fbox{
  \parbox{0.452\textwidth}{
  \textbf{Definition 2:} We define the \emph{fidelity} of explanation in IML as the faithfulness degree with regard to the target system, aiming to measure the relevance of explanations in practical settings.} 
}  \\

The third general property is \emph{persuasibility}, which reflects the degree of how human comprehend and response to the generated explanations. This property handles the subjective aspect of explanation, and is typically measured with human involvement. Good explanations are most likely to be easily comprehended, and facilitate quick responses from human users. Towards different user groups or application scenarios, one specific set of explanations could possibly have different persuasibility due to the diversified preferences. Thus, discussing persuasibility for explanation should only be considered under a same setting of users and tasks. \\  

\noindent
\fbox{
  \parbox{0.452\textwidth}{
  \textbf{Definition 3:} We define the \emph{persuasibility} of explanation in IML as the usefulness degree to human users, serving as the measure of subjective satisfaction or comprehensibility for the corresponding explanation.} 
}  \\

\subsection{Explanation Evaluation Problem}

\begin{figure}[t]
  \centering
  \includegraphics[width=\columnwidth]{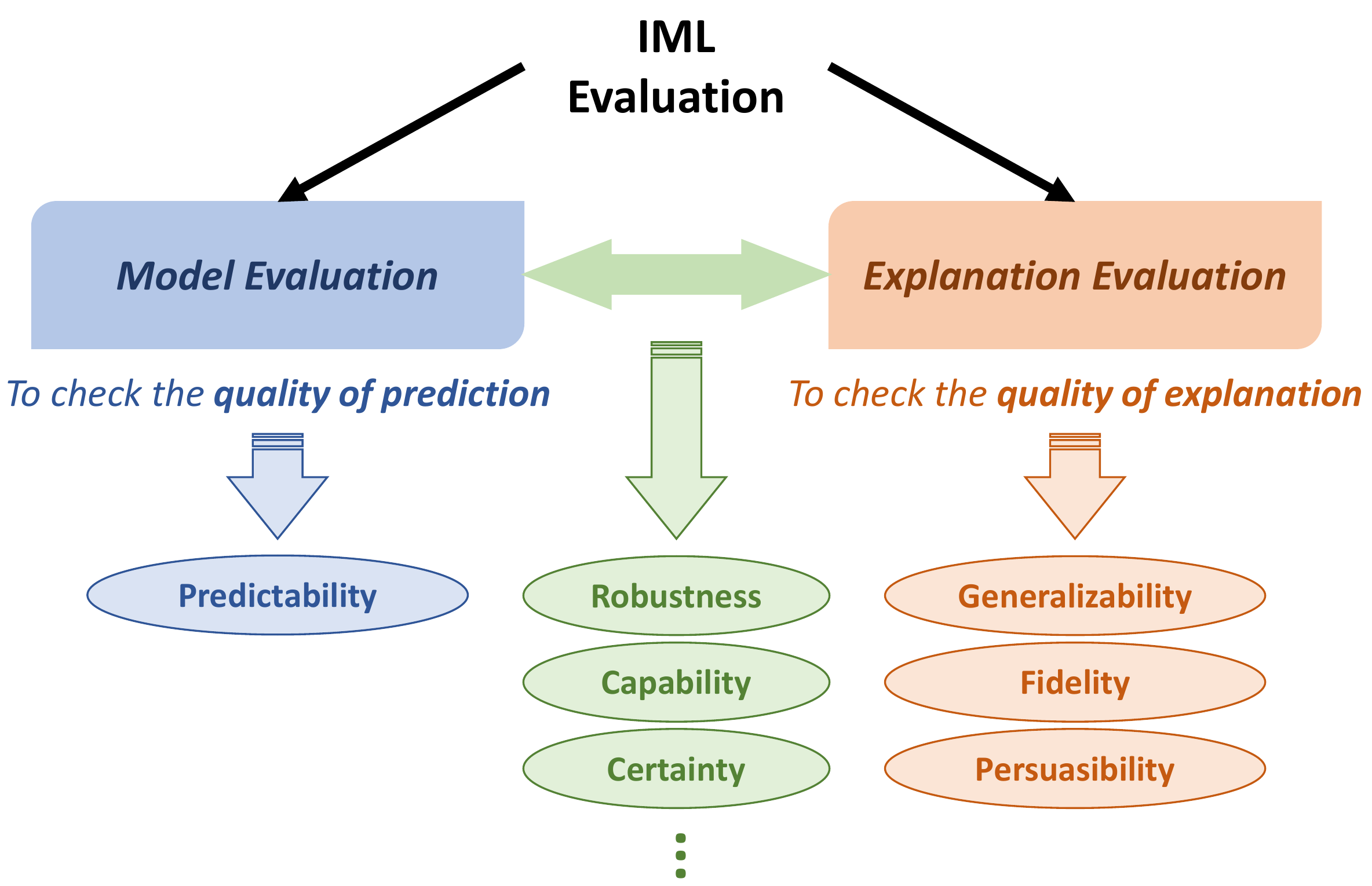}
  \caption{Illustration of the IML evaluation. Basically, IML evaluation can be divided into model evaluation and explanation evaluation. For model evaluation, we focus on the generalizability of the system, and evaluate the quality of prediction. For explanation evaluation, we focus on the predictability, fidelity, persuasibility, and evaluate the quality of explanation. Besides, there are also some special properties that are entangled with both model and explanation. We list robustness, capability and certainty here for instance. In this paper, we specifically focus on the aspects which are related to explanation evaluation.} 
  \label{fig:imleval}
\end{figure}

With the definitions of the three general properties for explanation, we further introduce and formally define the problem of evaluating explanations in IML. Technically, IML evaluation can be divided into two parts: \emph{model evaluation} and \emph{explanation evaluation}, shown by Figure~\ref{fig:imleval}. As for the model evaluation part, the goal is to measure the predictive power of IML systems, which is identical to that of common machine learning systems and can be directly achieved with some conventional metrics (e.g., accuracy, precision, recall and F1-score). Explanation evaluation, however, is different from model evaluation in both objective and methodology aspects. Since explanation typically contains more than one perspective and has no common ground truth over different scenarios, traditional model evaluation techniques thus cannot be perfectly applied. In this article, we specifically focus on the second part of IML evaluation, i.e., the explanation evaluation, and rigorously define the problem as follows.  \\ \\
\noindent
\fbox{
  \parbox{0.452\textwidth}{
  \textbf{Definition 4:} The \emph{explanation evaluation} problem within IML context is to assess the quality of the generated explanations from systems, where high-quality explanation corresponds to large values of generalizability, fidelity and persuasibility with relevant measurement. In general, good explanation ought to be \emph{well generalized}, \emph{highly faithful} and \emph{easily understood}. } 
}

\section{Evaluation Review}

In this section, we will conduct a systematic review for explanation evaluation problem in IML, following the properties of explanation we summarize. For each property, we mainly review the primary methodologies of evaluation for practical tasks, and shed light on the philosophy about why they are reasonable. After the review of evaluations on generalizability, fidelity and persuasibility, we also focus on some other special aspects, which are typically entangled together with model evaluation, including the \emph{robustness}, \emph{capability} and \emph{certainty}.

\subsection{Evaluation on Generalizability} 

Existing work, related to generalizability evaluation, mainly focus on the IML systems with intrinsic-global explanations. Since intrinsic-global explanations are typically presented and organized as the form of prediction models, it is straightforward and convenient to evaluate generalizability by applying those explanations on test data to see the corresponding generalization performance. Under this scenario, the generalizability evaluation task is somewhat equivalent to the model evaluation, where traditional model performance indicators (e.g., accuracy and F1-score) can be employed as the metrics to quantify the explanation generalizability. Conventional examples for this scenario can include generalized linear model (with informative coefficients)~\cite{nelder1972generalized}, decision tree (with structured branches)~\cite{safavian1991survey}, K-nearest neighbors (with significant instances)~\cite{altman1992introduction} and rule-based systems (with discriminative patterns)~\cite{buchanan1983principles}. In general, the generalizability evaluation for intrinsic-global explanations can be easily converted to model evaluation tasks, in which generalizability is positively correlated to the model prediction performance. Take the recent work on decision set~\cite{lakkaraju2016interpretable} for example. The authors use the AUC scores, a common metric for classification tasks in model evaluation, to indicate the generalizability of explanations in set of decision rules. Similarly in recent work~\cite{letham2015interpretable}, AUC scores are employed to reflect the explanation generalizability indirectly. 

Besides, there is another branch of work focusing on the generalizability of posthoc-global explanations. The relevant evaluation method is similar to that of generalizability for intrinsic-global ones, where model evaluation techniques could be employed to indicate the explanation generalizability. The major difference lies in the fact that the explanations we apply on test data are not directly associated with the target system, but are closely related to the interpretable proxies extracted or learned from the target. Those proxies typically serve as the role for interpreting the target system which is either a black box or a sophisticated model. Representative examples for this scenario can be found in knowledge distillation~\cite{frosst2017distilling} and mimic learning~\cite{che2015distilling}, where the common focus is to derive an interpretable proxy out of the black-box neural model for providing explanations. For example, in literature~\cite{che2015distilling}, the authors employ Gradient Boosting Trees (GBT) as the interpretable proxy to explain the working mechanism of neural networks. The constructed GBT is capable of providing feature importance for explanation, and is assessed by model evaluation techniques with AUC score to show the generalizability of corresponding explanations. The generalizability of posthoc-global explanation typically has positive correlation with the model performance of the derived interpretable proxy.

\subsection{Evaluation on Fidelity} 

Though pretty important for explanation evaluation, fidelity may not be explicitly considered for intrinsic explanations. In fact, the intrinsicality from explanations is sufficient to guarantee the exact working mechanism of the target IML system, and the corresponding explanations can thus be treated as faithful ones with full fidelity. The interpretable decision set~\cite{lakkaraju2016interpretable} should be a good example here. The learned decision set is self-interpretable and explicitly presents the decision rules for the potential classification tasks. Under this example, we can see that those explanation rules faithfully reflect the model prediction behaviour, and there does not exist any inconsistency between the IML system prediction and the relevant explanations. This kind of complete accordance between model and explanation is just what the full fidelity indicates. 

However, different from intrinsic ones, posthoc-global explanations in form of interpretable proxies cannot be regarded with full fidelity, since the derived proxies usually work in a different way compared with the target system. Although most proxies are derived to approximate the behaviour of target system, it is still constructed as a different model for the potential task. Existing work, related to fidelity evaluation for interpretable proxies, mainly use the difference in prediction performance to indicate the fidelity degree. For instance, in work~\cite{che2015distilling}, the authors conducted experiments with several sets of teacher-student models, where the teacher is the target model and the student is the proxy model. During the evaluation, the prediction differences between corresponding teachers and students are used to reflect the fidelity of the derived proxies, and preferred faithful proxies are shown to have minor losses in performance. 

Moreover, due to the posthoc manner and locality from nature, none of posthoc-local explanations is fully faithful to the target IML system. Among existing work, common ways to measure fidelity for posthoc-local explanations are ablation analysis~\cite{nguyen2015deep} and meaningful perturbations~\cite{fong2017interpretable}, where the core idea is to check the prediction variation after the adversarial changes made according to the generated explanations. The philosophy of this kind of methodologies is simple, i.e., modifications to the input instances, which are in accordance with the generated explanations, can bring about significant differences to model predictions if the explanations are faithful to the target system. Typical example can be found in image classification task with deep neural networks~\cite{selvaraju2017grad}, where the fidelity of generated posthoc-local explanations are evaluated by measuring the prediction difference between the original image and the perturbative image. The overall logic here is to mask the attributing regions in images indicated by the explanations, and then check the extent of prediction variation. The larger the difference, the more faithful the generated explanations are. In addition to the image classification task, ablation- and perturbation-based fidelity evaluation methods have also been effectively used in text classification~\cite{du2019attribution}, recommender system~\cite{yang2018towards} and adversarial detection~\cite{liu2018adversarial}. Furthermore, as for the posthoc-local explanations in form of training samples~\cite{koh2017understanding} and model components~\cite{narendra2018explaining}, ablation and perturbation operations are properly applied as well in evaluating the explanation fidelity.

\subsection{Evaluation on Persuasibility} 

To effectively evaluate the persuasibility of generated explanations, human annotation is widely used especially in those uncontentious tasks, such as object detection. The annotation-based evaluation is usually regarded to be objective, since relevant annotations do not change among different groups of user. In computer vision related tasks, the most common annotations for persuasibility evaluation are bounding box~\cite{szegedy2013deep} and semantic segmentation~\cite{long2015fully}. Appropriate example can be found in recent work~\cite{selvaraju2017grad}, which utilizes bounding boxes to evaluate the persuasibility of explanations and employ the metric Intersection over Union (IoU) or Jaccard index to quantify the persuasibility performance. As for the annotations with semantic segmentation, recent work~\cite{zhou2018interpreting} employs the pixel-level difference as the metric to measure the corresponding persuasibility of explanations. Moreover, in natural language processing, similar human annotation, named rationale~\cite{lei2016rationalizing}, has been extensively used for evaluation, which is a subset of features highlighted by annotators and regarded to be important for prediction. Through those different forms of annotations, the persuasibility of explanation can be objectively evaluated with human-grounded truth, which typically keeps consistent across different groups of user and one particular task. Due to the one-to-one correspondence between the annotation and the instance, annotation-based evaluation is usually applied to those local explanations instead of the global ones. 

Conducting persuasibility evaluation with human annotation does not work well in complicated tasks, since the related annotations may not keep consistent across different user groups. Under those circumstances, employing users for human studies is the common way to evaluate the persuasibility of explanation. To appropriately design relevant human studies, both machine learning experts and human-computer interaction (HCI) researchers actively explore this area~\cite{abdul2018trends,herman2017promise}, and propose several metrics for human evaluation on general explanation from IML techniques, such as mental mode~\cite{lakkaraju2016interpretable}, human-machine performance~\cite{feng2018can}, user satisfaction~\cite{lage2019evaluation} and user trust~\cite{holliday2016user}. Take the most recent work~\cite{lage2019evaluation} for instance. The authors focus on the user satisfaction in evaluating the persuasibility, and specifically employ the human response time and decision accuracy as the auxiliary metrics. The whole study is conducted on two different domains with three types of explanation variation, aiming to conclude the relation between the explanation quality and human cognition. With the aid of human studies, persuasibility of explanation can be evaluated under a more complicated and practical setting, regarding to specific user groups and application domains. By directly measuring explanations from human users, we can realize the usefulness in real-world applications when determining the explanation quality. Since human studies can be designed flexibly according to diversified needs and goals, this methodology is generally applicable to all kinds of explanations for persuasibility evaluation within IML context.

\subsection{Evaluation on Other Properties} 

Besides the generalizability, fidelity and persuasibility, existing work also consider some other properties when evaluating the explanation in IML. We introduce those properties separately due to the following two reasons. First, those properties are not representative and general for explanation evaluation among IML systems, and are simply considered under specific architectures or applications. Second, those properties are related to both prediction model and generated explanation, which typically need novel and special design to evaluate. In this part, we particularly focus on the following three properties. 

\emph{Robustness.} 
Similar to machine learning models, the generated explanations from IML systems can also be fragile to adversarial perturbations, especially for those posthoc ones from neural architectures~\cite{ghorbani2017interpretation}. Explanation robustness is primarily designed to measure how similar the explanations are for similar instances. Recent work~\cite{selvaraju2017grad,yeh2019sensitive} all conduct robustness evaluation for explanation with the metrics on sensitivity, beyond the evaluation on those three general properties we summarize. Robust explanations are always preferred in building a trustable IML system for human users. To obtain the explanations with high robustness, a stable prediction model and a reliable explanation generation algorithm are usually the two most important keys.

\emph{Capability.} 
Another property for explanation evaluation is named capability, which is used to indicate the extent that corresponding explanations can be generated. Commonly, this property is evaluated on those explanations generated from search based methodologies~\cite{wallace2018interpreting}, instead of those obtained from gradient based~\cite{selvaraju2017grad} or perturbation based~\cite{ribeiro2016should} methodologies. Typical example for capability evaluation can be found in work~\cite{yang2018towards} with the application to recommender system, where the authors employ the explainability precision and explainability recall as the metrics to indicate the capability strength. Similar to the property robustness, capability is also related to the target prediction model, which essentially determines the upper bound of the ability to generate explanations.

\emph{Certainty.} 
To further evaluate explanations on whether they reflect the uncertainty of the target IML system, existing work also focus on the certainty aspect of explanation. Certainty is also a property related to both model and explanation, since explanation can only provide uncertainty interpretation as long as the corresponding IML system itself has the certainty measure. Recent work~\cite{Phillips2017InterpretableAL} gives an appropriate example for certainty evaluation. In this work, the authors consider the IML systems under the active learning settings, and propose a novel measure, named uncertainty bias, to evaluate the certainty of generated explanations. Specifically, the explanation certainty is measured according to the discrepancy in prediction confidence of the IML system between one category and the others. In similar ways, work~\cite{ustun2019actionable} focus on the certainty aspect of explanations as well, and provide insights on how confident users could be for particular outputs with the computed explanations in form of flip set (i.e., a list of actionable changes that users can make to flip the prediction of the target system). In essence, certainty evaluation and persuasibility evaluation can be mutually supported from each other.

\section{Discussion and Exploration}

In this section, we first propose a unified framework to conduct general assessment on explanations in IML, according to the different level of needs for evaluation. Then, several open problems in explanation evaluation are raised and discussed regarding to benchmarking issues. Further, we highlight some significant limitations of current evaluation techniques for future exploration.

\subsection{Unified Framework for Evaluation} 

Despite the large number of work we reviewed for explanation evaluation, different work typically have their own particular focus, depending on the specific tasks, architectures, or applications. This situation leads to the fact that it is hard to benchmark the evaluation process for explanations in IML as what we developed in model evaluation. To pave the way to benchmark evaluation on explanation, we try to construct a unified framework here by considering those properties of explanations. To make the framework general, we simply take the generalizability, fidelity and persuasibility into account, and do not consider those special ones under particular scenarios.

\subsubsection{Different level of needs for evaluation} 

Although we conduct the review separately, regarding to generalizability, fidelity and persuasibility, those three general properties are internally related to each other, where each of them represents a specific level of needs for evaluation. From the lower level to higher level, we can sort the properties as: generalizability, fidelity, persuasibility. Generalizability typically serves as the basic need in evaluation, since it formulates the foundation for other properties. In real-world applications, good generalizability is the precondition for human users to make accurate decisions with the generated explanations, which guarantees that the explanations we employ are generalizable and reflect the true knowledge for particular tasks. After that, a further demand for human users is to check whether the derived explanations at hands are reliable or not. This demand pushes out the fidelity property to the front. By assessing the fidelity, better decisions can be made on whether to trust the IML system or not based on the explanation relevance. As for the higher demand on real effectiveness in practice, persuasibility is further considered to indicate the tangible impacts, directly bridging the gap between human users and machine explanations. For one specific task, the explanation evaluation mainly depends on the corresponding applications and user groups, which determine the level of needs in evaluation design. Generally, model developers would care more on those basic properties of lower levels, including generalizability and fidelity, while general end-users would pay more attention on the persuasibility in a higher level. 

\subsubsection{Hierarchical structure of the framework} 

\begin{figure}[t]
  \vspace{5pt}
  \centering
  \includegraphics[width=\columnwidth]{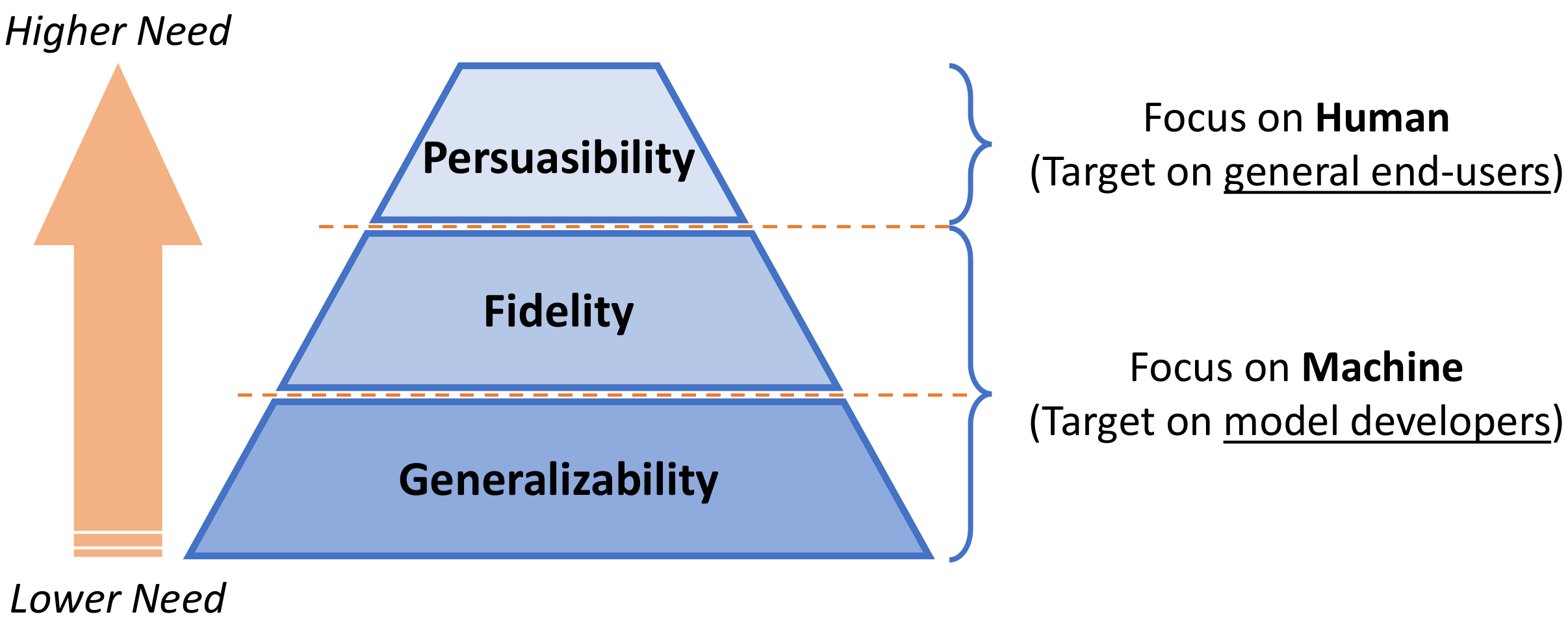}
  \caption{A unified hierarchical framework for explanation evaluation in IML. The whole framework consists of three different tiers, corresponding to generalizability, fidelity and persuasibility, from the lower level to the higher level. Basically, the bottom and medium tier focus on the evaluation from machine perspective, while the top tier concentrate on the evaluation from human perspective. To this end, the bottom and medium tier are usually designed for model developers, and the top tier is designed for general end-users.}
  \label{fig:evalframe}
\end{figure}

The overall unified evaluation framework is designed hierarchically, according to the different level of needs, as illustrated in Figure~\ref{fig:evalframe}. In the bottom tier, the evaluation goal focuses on the generalizability, where generated explanations are tested for their generalization power. In the medium tier, the goal is to evaluate the fidelity, with regard to the target IML system. The top tier aims to evaluate the persuasibility, targeting on specific applications and user groups. To have a unified evaluation in one particular task, each tier should have a consistent pipeline with a fixed set of data, user and metrics correspondingly. The overall evaluation results can be further derived through an ensemble way, such as weighted sum, where each tier could be assigned with an importance weight depending on the applications and user groups. This proposed hierarchical framework is generally applicable to most of explanation evaluation problems, which could be appended with new components if necessary. With proper metrics, as well as a sensible manner for ensemble, the framework can effectively help human users measure the overall quality of explanation from IML techniques under certain circumstances. 

\subsection{Open Problems for Benchmark} 

To fully achieve the benchmark for explanation evaluation in real-world applications, there are still some open problems left to explore, which are listed and discussed as follows. 

\subsubsection{Generalizability for local explanations}   

Existing work on generalizability evaluation mainly focus on those global explanations, while limited efforts has been paid on the local ones. The challenges in evaluating generalizability of local explanations are in two folds: (1) local explanations cannot be easily organized into valid prediction models, which makes the model evaluation techniques hard to be directly applied; (2) local explanations simply contain the partial knowledge learned by the target IML system, thus special designs are required to ensure the evaluation has a specific local focus. Though no direct solutions, some insights from existing efforts may be inspiring. As for the first challenge, an approximated local classifier~\cite{ribeiro2016should} could be potentially built to carry the local explanations, and then the generalizability could be further assessed with model evaluation techniques by specifying test instances. Moreover, for the second challenge, we could possibly employ local explanations, together with human simulated/augmented data, to train a separate classifier~\cite{kim2017interpretability} for generalizability evaluation, where the task is essentially reduced from the original one and only involves the local knowledge we test with. 

\subsubsection{Fidelity for posthoc explanations} 
Among existing work, it is well received that good explanation should have high fidelity to the target IML system. However, with the posthoc manner, it might not be the case that faithful explanations are always the good ones that human user prefer. During explanation evaluation, we typically assume that IML systems are well trained and are capable of making reasonable decisions, but this assumption is hard to be perfectly achieved in practice. As a result, the generated post-hoc explanations may not be with high quality due to the inferior model performance, although they might be highly faithful to the target system. Thus, designing a novel methodology, which could consider both model and explanation, for posthoc fidelity evaluation is of great importance. In general, how to utilize the model performance to guide the measurement of posthoc explanation fidelity is the key problem to tackle this challenge, where the ultimate goal is to help human users better select out those explanations with good quality from fidelity perspective. 

\subsubsection{Persuasibility for global explanations} 

As for the persuasibility, it is also challenging to conduct effective evaluations on global explanations, no matter using annotation based methods or employing human studies. The main reason lies in the fact that global explanations in real applications are very sophisticated, which makes it hard to make annotations or select appropriate users for studies. Essentially, the global nature requires either selected annotators or users to equip with comprehensive understandings towards the target task, otherwise the evaluation results would be less convincing or even misleading. Besides, the global explanations in practice typically contain tons of information, which could be extremely time-consuming to evaluate persuasibility. One possible solution is to use some simplified or proxy tasks to simulate the original one, as mentioned in ~\cite{doshi2017towards}, but this kind of substitution needs to maintain the original essence, which certainly requires non-trivial efforts on task abstraction. Another potential solution is to simplify the explanations shown to users, such as only showing the top-k features, which, however, sacrifices the comprehensiveness of generated explanations and impedes the full view over the target system.

\subsection{Limitations of Current Evaluation} 

Although various methodologies of explanation evaluation exist in IML research, there are still some significant limitations of current evaluation techniques. We briefly introduce some of the most important ones as below. 

\subsubsection{Causality insight for evaluation}

The first limitation lies in the \emph{lack of causal perspective}~\cite{kim2019learning} in explanation evaluation. Current evaluation techniques, no matter what properties they focus on, mostly fail to have causal analysis when evaluating the explanation quality. This kind of drawback could possibly lead to the fact that our selected explanations may not fully represent the true reasons behind the prediction, since the influence from confounders are not effectively blocked during interpretation. Take the two most common methodologies in IML, gradient based and perturbation based methods, for examples. Both of them can be viewed as special cases of Individual Causal Effect (ICE) analysis, where complicated inter-feature interactions could conceal the real importance of some input features~\cite{chattopadhyay19a}. Thus, to derive better explanations with relevant causal guarantees, we need corresponding evaluation techniques to assess the causal perspective of the generated explanations. In this way, human users would be further enabled to have a clearer understanding towards the cause-effect association when interpreting the target system.

\subsubsection{Completeness insight for evaluation}

The second limitation is the \emph{neglect of completeness} in explanation evaluation~\cite{gilpin2018explaining}. Existing efforts on IML evaluation cannot well reflect the degree of completeness for generated explanations, which makes it difficult for human users to further ensure the real value in practice. Explanation completeness could be important in real applications, because it is able to indicate the possibility of whether there would be additional explanations for certain prediction results. Questions, such as ``\textit{Do we get the full explanations from the target IML system?}" and ``\textit{Is it possible to generate better explanations than the current ones?}", are not supported by the current evaluation techniques. A completeness-aware evaluation for explanation would definitely be helpful in exploring the boundaries of the target IML system. Besides, having completeness insight for assessment would also be a significant supplement for persuasibility evaluation, since the need for explainability typically stems from the incompleteness in problem formalization~\cite{doshi2017towards}.

\subsubsection{Novelty insight for evaluation}

The third limitation results from the \emph{explanation novelty perspective}~\cite{preuer2019interpretable}. Under the current infrastructure of explanation evaluation in IML, it is commonly assumed that high-quality explanations are those ones which can help human users make better decisions or obtain better understandings. Nevertheless, the view of this assumption for good explanation is rather limited, since it somewhat overlooks the potential values of the explanations that may not be well comprehended by users. Explanations which are not directly ``useful'' to human users may still have significant influences, due to their important roles in extending the human knowledge boundary. Medical diagnosis should be a good example to illustrate this point. When diagnosing patients, doctors would typically refer the generated explanations with their acquired domain knowledge, if they have access to the IML systems. Since there is no way that domain knowledge can cover all aspects and contain full pathological mechanism, especially for those new diseases, we cannot casually discard the explanations that are mismatched with our knowledge. Those ``novel'' explanations could possibly point out some valuable research areas in a reverse way. To this end, current evaluation techniques need to be further enhanced to properly cover the novelty issue in assessing the quality of generated explanations, so that novel explanations could be well distinguished from those noisy ones.

\section{Conclusions}

With the booming development of IML techniques, how to effectively evaluate those generated explanations, typically without ground truth on quality, is becoming increasingly critical in recent years. In this article, we briefly introduce the explanation in IML, as well as its three general properties, and formally define the explanation evaluation problem within the context of IML. Then, following the properties, we systematically review the existing efforts in evaluating explanation, covering various methodologies and application scenarios. Moreover, a potential unified evaluation framework is built according to the hierarchical needs from both model developers and general end-users. In the end, several open problems in benchmark and limitations of current techniques are discussed for future exploration. Though numerous obstacles are still left to be solved, explanation evaluation will keep playing the key role in enabling effective interpretation of IML systems.


\balance

\bibliographystyle{abbrv} 
\bibliography{bibs4cacm}

\begin{thebibliography}{10}

\bibitem{abdul2018trends}
A.~Abdul, J.~Vermeulen, D.~Wang, B.~Y. Lim, and M.~Kankanhalli.
\newblock Trends and trajectories for explainable, accountable and intelligible
  systems: An hci research agenda.
\newblock In {\em Proceedings of the 2018 CHI Conference on Human Factors in
  Computing Systems}, page 582. ACM, 2018.

\bibitem{altman1992introduction}
N.~S. Altman.
\newblock An introduction to kernel and nearest-neighbor nonparametric
  regression.
\newblock {\em The American Statistician}, 46(3):175--185, 1992.

\bibitem{buchanan1983principles}
B.~G. Buchanan and R.~O. Duda.
\newblock Principles of rule-based expert systems.
\newblock In {\em Advances in computers}, volume~22, pages 163--216. Elsevier,
  1983.

\bibitem{chattopadhyay19a}
A.~Chattopadhyay, P.~Manupriya, A.~Sarkar, and V.~N. Balasubramanian.
\newblock Neural network attributions: A causal perspective.
\newblock In {\em Proceedings of the 36th International Conference on Machine
  Learning}, volume~97, pages 981--990, 2019.

\bibitem{che2015distilling}
Z.~Che, S.~Purushotham, R.~Khemani, and Y.~Liu.
\newblock Distilling knowledge from deep networks with applications to
  healthcare domain.
\newblock {\em arXiv preprint arXiv:1512.03542}, 2015.

\bibitem{doshi2017towards}
F.~Doshi-Velez and B.~Kim.
\newblock Towards a rigorous science of interpretable machine learning.
\newblock {\em arXiv preprint arXiv:1702.08608}, 2017.

\bibitem{du2018techniques}
M.~Du, N.~Liu, and X.~Hu.
\newblock Techniques for interpretable machine learning.
\newblock {\em Communications of the ACM}, 2019.

\bibitem{du2019attribution}
M.~Du, N.~Liu, F.~Yang, S.~Ji, and X.~Hu.
\newblock On attribution of recurrent neural network predictions via additive
  decomposition.
\newblock In {\em Proceedings of The Web Conference 2019 (TheWebConf)}. ACM,
  2019.

\bibitem{feng2018can}
S.~Feng and J.~Boyd-Graber.
\newblock What can ai do for me: Evaluating machine learning interpretations in
  cooperative play.
\newblock {\em arXiv preprint arXiv:1810.09648}, 2018.

\bibitem{fong2017interpretable}
R.~C. Fong and A.~Vedaldi.
\newblock Interpretable explanations of black boxes by meaningful perturbation.
\newblock In {\em Proceedings of the IEEE International Conference on Computer
  Vision}, pages 3429--3437, 2017.

\bibitem{frosst2017distilling}
N.~Frosst and G.~Hinton.
\newblock Distilling a neural network into a soft decision tree.
\newblock {\em arXiv preprint arXiv:1711.09784}, 2017.

\bibitem{ghorbani2017interpretation}
A.~Ghorbani, A.~Abid, and J.~Zou.
\newblock Interpretation of neural networks is fragile.
\newblock {\em arXiv preprint arXiv:1710.10547}, 2017.

\bibitem{gilpin2018explaining}
L.~H. Gilpin, D.~Bau, B.~Z. Yuan, A.~Bajwa, M.~Specter, and L.~Kagal.
\newblock Explaining explanations: An overview of interpretability of machine
  learning.
\newblock In {\em 2018 IEEE 5th International Conference on Data Science and
  Advanced Analytics (DSAA)}, pages 80--89. IEEE, 2018.

\bibitem{herman2017promise}
B.~Herman.
\newblock The promise and peril of human evaluation for model interpretability.
\newblock {\em arXiv preprint arXiv:1711.07414}, 2017.

\bibitem{holliday2016user}
D.~Holliday, S.~Wilson, and S.~Stumpf.
\newblock User trust in intelligent systems: A journey over time.
\newblock In {\em Proceedings of the 21st International Conference on
  Intelligent User Interfaces}, pages 164--168. ACM, 2016.

\bibitem{kim2017interpretability}
B.~Kim, M.~Wattenberg, J.~Gilmer, C.~Cai, J.~Wexler, F.~Viegas, and R.~Sayres.
\newblock Interpretability beyond feature attribution: Quantitative testing
  with concept activation vectors (tcav).
\newblock {\em arXiv preprint arXiv:1711.11279}, 2017.

\bibitem{kim2019learning}
C.~Kim and O.~Bastani.
\newblock Learning interpretable models with causal guarantees.
\newblock {\em arXiv preprint arXiv:1901.08576}, 2019.

\bibitem{koh2017understanding}
P.~W. Koh and P.~Liang.
\newblock Understanding black-box predictions via influence functions.
\newblock In {\em Proceedings of the 34th International Conference on Machine
  Learning-Volume 70}, pages 1885--1894. JMLR, 2017.

\bibitem{lage2019evaluation}
I.~Lage, E.~Chen, J.~He, M.~Narayanan, B.~Kim, S.~Gershman, and F.~Doshi-Velez.
\newblock An evaluation of the human-interpretability of explanation.
\newblock {\em arXiv preprint arXiv:1902.00006}, 2019.

\bibitem{lakkaraju2016interpretable}
H.~Lakkaraju, S.~H. Bach, and J.~Leskovec.
\newblock Interpretable decision sets: A joint framework for description and
  prediction.
\newblock In {\em Proceedings of the 22nd ACM SIGKDD international conference
  on knowledge discovery and data mining}, pages 1675--1684. ACM, 2016.

\bibitem{lecun2015deep}
Y.~LeCun, Y.~Bengio, and G.~Hinton.
\newblock Deep learning.
\newblock {\em Nature}, 521(7553):436, 2015.

\bibitem{lei2016rationalizing}
T.~Lei, R.~Barzilay, and T.~Jaakkola.
\newblock Rationalizing neural predictions.
\newblock In {\em Proceedings of the Conference on Empirical Methods in Natural
  Language Processing}. NIH Public Access, 2016.

\bibitem{letham2015interpretable}
B.~Letham, C.~Rudin, T.~H. McCormick, and D.~Madigan.
\newblock Interpretable classifiers using rules and bayesian analysis: Building
  a better stroke prediction model.
\newblock {\em The Annals of Applied Statistics}, 9(3):1350--1371, 2015.

\bibitem{liu2018adversarial}
N.~Liu, H.~Yang, and X.~Hu.
\newblock Adversarial detection with model interpretation.
\newblock In {\em Proceedings of the 24th ACM SIGKDD International Conference
  on Knowledge Discovery \& Data Mining}, pages 1803--1811. ACM, 2018.

\bibitem{long2015fully}
J.~Long, E.~Shelhamer, and T.~Darrell.
\newblock Fully convolutional networks for semantic segmentation.
\newblock In {\em Proceedings of the IEEE conference on computer vision and
  pattern recognition}, pages 3431--3440, 2015.

\bibitem{narendra2018explaining}
T.~Narendra, A.~Sankaran, D.~Vijaykeerthy, and S.~Mani.
\newblock Explaining deep learning models using causal inference.
\newblock {\em arXiv preprint arXiv:1811.04376}, 2018.

\bibitem{nelder1972generalized}
J.~A. Nelder and R.~W. Wedderburn.
\newblock Generalized linear models.
\newblock {\em Journal of the Royal Statistical Society: Series A (General)},
  135(3):370--384, 1972.

\bibitem{nguyen2015deep}
A.~Nguyen, J.~Yosinski, and J.~Clune.
\newblock Deep neural networks are easily fooled: High confidence predictions
  for unrecognizable images.
\newblock In {\em Proceedings of the IEEE conference on computer vision and
  pattern recognition}, pages 427--436, 2015.

\bibitem{Phillips2017InterpretableAL}
R.~L. Phillips, K.~H. Chang, and S.~A. Friedler.
\newblock Interpretable active learning.
\newblock In {\em FAT}, 2017.

\bibitem{preuer2019interpretable}
K.~Preuer, G.~Klambauer, F.~Rippmann, S.~Hochreiter, and T.~Unterthiner.
\newblock Interpretable deep learning in drug discovery.
\newblock {\em arXiv preprint arXiv:1903.02788}, 2019.

\bibitem{ribeiro2016should}
M.~T. Ribeiro, S.~Singh, and C.~Guestrin.
\newblock Why should i trust you?: Explaining the predictions of any
  classifier.
\newblock In {\em Proceedings of the 22nd ACM SIGKDD international conference
  on knowledge discovery and data mining}, pages 1135--1144. ACM, 2016.

\bibitem{safavian1991survey}
S.~R. Safavian and D.~Landgrebe.
\newblock A survey of decision tree classifier methodology.
\newblock {\em IEEE transactions on systems, man, and cybernetics},
  21(3):660--674, 1991.

\bibitem{selvaraju2017grad}
R.~R. Selvaraju, M.~Cogswell, A.~Das, R.~Vedantam, D.~Parikh, and D.~Batra.
\newblock Grad-cam: Visual explanations from deep networks via gradient-based
  localization.
\newblock In {\em Proceedings of the IEEE International Conference on Computer
  Vision}, pages 618--626, 2017.

\bibitem{szegedy2013deep}
C.~Szegedy, A.~Toshev, and D.~Erhan.
\newblock Deep neural networks for object detection.
\newblock In {\em Advances in neural information processing systems}, pages
  2553--2561, 2013.

\bibitem{ustun2019actionable}
B.~Ustun, A.~Spangher, and Y.~Liu.
\newblock Actionable recourse in linear classification.
\newblock In {\em Proceedings of the Conference on Fairness, Accountability,
  and Transparency}, pages 10--19. ACM, 2019.

\bibitem{wachter2017transparent}
S.~Wachter, B.~Mittelstadt, and L.~Floridi.
\newblock Transparent, explainable, and accountable ai for robotics.
\newblock {\em Science Robotics}, 2(6), 2017.

\bibitem{wallace2018interpreting}
E.~Wallace, S.~Feng, and J.~Boyd-Graber.
\newblock Interpreting neural networks with nearest neighbors.
\newblock {\em arXiv preprint arXiv:1809.02847}, 2018.

\bibitem{yang2018towards}
F.~Yang, N.~Liu, S.~Wang, and X.~Hu.
\newblock Towards interpretation of recommender systems with sorted explanation
  paths.
\newblock In {\em 2018 IEEE International Conference on Data Mining (ICDM)},
  pages 667--676. IEEE, 2018.

\bibitem{yeh2019sensitive}
C.-K. Yeh, C.-Y. Hsieh, A.~S. Suggala, D.~Inouye, and P.~Ravikumar.
\newblock How sensitive are sensitivity-based explanations?
\newblock {\em arXiv preprint arXiv:1901.09392}, 2019.

\bibitem{zhou2018interpreting}
B.~Zhou, D.~Bau, A.~Oliva, and A.~Torralba.
\newblock Interpreting deep visual representations via network dissection.
\newblock {\em IEEE transactions on pattern analysis and machine intelligence},
  2018.

\end{thebibliography}

\end{document}